\documentclass[conference]{IEEEtran}
\IEEEoverridecommandlockouts
\usepackage{amsmath, amssymb, amsfonts, nccmath, mathrsfs}
\usepackage{algorithm}
\usepackage{algpseudocode}
\usepackage{array}
\usepackage{booktabs}
\usepackage{cite}
\usepackage{dsfont}
\usepackage{graphicx}
\usepackage{hyperref}
\usepackage{longtable}  
\usepackage{lipsum}
\usepackage{multirow}
\usepackage{pifont}
\usepackage{subfigure}
\usepackage{soul}
\usepackage{svg}
\usepackage{textcomp}
\usepackage[normalem]{ulem}
\usepackage{xcolor}

\newcolumntype{C}[1]{>{\centering\arraybackslash}m{#1}}

\begin{document}

\title{A Method for Discovering Novel Classes \\in Tabular Data
}

\author{
    \IEEEauthorblockN{
        Colin Troisemaine\IEEEauthorrefmark{1}\IEEEauthorrefmark{2},
        Joachim Flocon-Cholet\IEEEauthorrefmark{1},
        Stéphane Gosselin\IEEEauthorrefmark{1}, \\
        Sandrine Vaton\IEEEauthorrefmark{2},
        Alexandre Reiffers-Masson\IEEEauthorrefmark{2},
        Vincent Lemaire\IEEEauthorrefmark{1}
    }
    \IEEEauthorblockA{
        \IEEEauthorrefmark{1} \textit{Orange Labs, Lannion, France}\\
        \IEEEauthorrefmark{2} \textit{Department of Computer Science, IMT Atlantique, Brest, France}\\
        \textit{colin.troisemaine@orange.com}
    }
}

\maketitle

\begin{abstract}
In Novel Class Discovery (NCD), the goal is to find new classes in an unlabeled set given a labeled set of known but different classes.
While NCD has recently gained attention from the community, no framework has yet been proposed for heterogeneous tabular data, despite being a very common representation of data.
In this paper, we propose TabularNCD, a new method for discovering novel classes in tabular data.
%
%
We show a way to extract knowledge from already known classes to guide the discovery process of novel classes in the context of tabular data which contains heterogeneous variables.
A part of this process is done by a new method for defining pseudo labels, and we follow recent findings in Multi-Task Learning to optimize a joint objective function.
Our method demonstrates that NCD is not only applicable to images but also to heterogeneous tabular data.
%
%
Extensive experiments are conducted to evaluate our method and demonstrate its effectiveness against 3 competitors on 7 diverse public classification datasets.
%

\end{abstract}


\section{Introduction}
\label{sec:intro}

The recent success of machine learning models has been enabled in part by the use of large quantities of labeled data.
Many methods currently assume that a large part of the available data is labeled and that all the classes are known.
However, these assumptions don't always hold true in practice and researchers have started considering scenarios where unlabeled data is available \cite{NodetLBCO21a, Zhou2017}.
They belong to Weakly Supervised Learning, where methods that require all the classes to be known in advance can be distinguished from those that are able to manage classes that have never appeared during training.
As an example, Semi-Supervised Learning \cite{ChapelleSemi} combines a small amount of labeled data with larger amounts of unlabeled data.
Nonetheless, some labeled data is still needed for each of the classes in this case. Another example is Zero-Shot Learning \cite{10.1145/3293318}, where the models are designed to accurately predict classes that have never appeared during training. But some kind of description of these unknown classes is needed to be able to recognize them.

Very recently, Novel Class Discovery (NCD) \cite{hsu2018learning} has been proposed to fill these gaps and attempts to identify new classes in unlabeled data by exploiting prior knowledge from known classes. In this specific setup, the data is split in two sets. The first is a labeled set containing known classes and the second is an unlabeled set containing unknown classes that must be discovered. Some solutions have been proposed for the NCD problem in the context of computer vision \cite{autonovel2021, han2019learning, wang2020openworld, zhong2021neighborhood} and have displayed promising results. 

However, research in this domain is still recent, and to the best of our knowledge, NCD has not directly been addressed for tabular data. While audio and image data have been of great interest in recent scientific publications, tabular data remains a very common information structure that is found in many real world problems, such as the information systems of companies. In this paper, we will therefore focus on NCD for tabular data.

Tabular data can come in large unlabeled quantities due to the labelling process being often costly and time-consuming, in part due to the difficulty of visualization, which is why a lot of unlabeled data is often left unexploited. In \cite{deeptabularsurvey}, the authors review the primary challenges that come with tabular data and identify three main obstacles to the success of deep neural networks on this type of data. The first is the quality of the data, that often includes missing values, extreme data (outliers), erroneous or inconsistent data and class imbalance. Then, the lack of spatial correlation between features makes it difficult to use techniques based on inductive biases, such as convolutions or data augmentation \cite{tabularimages}. And finally, the heterogeneous and complex nature of the data (dense continuous and sparse categorical features) that can require considerable pre-processing, leading to information loss compared to the original data \cite{impactencoding}.

\textbf{Our proposal:} To address the NCD problem in the challenging tabular data environment, we propose TabularNCD (for Tabular Novel Class Discovery). In the first step of our method, an unbiased latent representation is initialized by taking advantage of the advances in Self-Supervised Learning (SSL) for tabular data \cite{NEURIPS2020_7d97667a}. Then, we build on the idea that the local neighborhood of an instance in the latent space is likely to belong to the same class, and a clustering of the unlabeled data is learned through similarity measures. The process in this second step is jointly optimized with a classifier on the known classes to include the relevant features from the already discovered classes.

Our key contributions are summarized as follows:
\begin{itemize}
    \item We propose TabularNCD, a new method for Novel Class Discovery. To the best of our knowledge, this is the first attempt at solving the NCD problem in the context of tabular data with heterogeneous features. Thus, we do not depend on the spatial inductive bias of features, which other NCD methods rely heavily on when using convolutions and specialized data augmentation techniques.
    \item We empirically evaluate TabularNCD on seven varied public classification datasets. These experiments demonstrate the superior performance of our method over common fully unsupervised methods and a baseline that exploits known classes in a naive way.
    \item We also introduce an original approach for the definition of pairs of pseudo labels of unlabeled data. This approach exploits the local neighborhood of an instance in a pre-trained latent space by considering that its $k$ most similar instances belong to the same class. We study its robustness and the influence of its parameters on performance.
    \item Lastly, we conduct experiments to understand in depth the reasons for the advantage that the proposed method has over simpler approaches.
\end{itemize}


\section{Novel Class Discovery and related works}
\label{sec:related_works}

The process of discovering new classes depends on the final application in the real world (offline vs online learning, nature of the data, etc). In the literature of ``concept drift'' \cite{Lu2019LearningUC}, changes in the distribution of known classes can appear either gradually or suddenly. Some analogies can be made with the present paper, where we consider the case of the ``sudden drift'', as the labeled dataset $D^l$ and unlabeled dataset $D^u$ do not share any classes (as described in Sec.~\ref{sec:intro}). Furthermore, we assume that a form of knowledge can be extracted from $D^l$ to be then used on $D^u$. In this sense, ``Novel Class Discovery (NCD)'' methods mainly lie at the intersection of several other lines of research that we briefly review here.


\textbf{Transfer Learning \cite{5288526}:} To solve a problem faster or with better generalization, transfer learning leverages knowledge from a different (but related) problem. Transfer learning can either be \textit{cross-domain}, when a model trained to execute a task on one domain is retrained to solve the same task but on another domain. Or it can be \textit{cross-task}, when a model trained to recognize some classes is retrained to distinguish other classes of the same domain.
NCD can be regarded as a cross-task transfer learning problem, in the sense that it aims to cluster unlabeled data by leveraging knowledge from different labeled data.
However, most of the works in transfer learning require labeled data in the source and target domains to train a classifier for the new task.
When there are no labels in the target domain, cross-task transfer learning methods usually employ unsupervised clustering approaches trained on features extracted from labeled data.

\textbf{Open World Learning:} Unlike in the traditional \textit{closed-world learning}, the problem of Open World Learning expects at test time instances from classes that were not seen during training. The objective is therefore threefold: 1) to obtain good accuracy on the known classes, 2) to recognize instances from novel classes and optionally, 3) to incrementally learn the new classes. In practice, most methods \cite{qin2020text, 2019HuXu} assume that the discovery of the new classes in the rejected examples is either the duty of a human or a task that is outside the scope of their research. To the best of our knowledge, \cite{2019Guo, shu2018unseen} are exceptions in this regard. However, Open World Learning still diverges from NCD, as classes in the unlabeled set can be unknown or not. This is not the case for NCD, where it is assumed that it is already known if an instance belongs to a unknown class or not. In other words, the focus of NCD is only on the third task of Open World Learning.

\textbf{Semi-Supervised Learning:} When a training set is partially labeled, classical supervised methods can only exploit the labeled data. Semi Supervised Learning \cite{ChapelleSemi} is a special instance of \textit{weak supervision} where additional information in the form of a labeled subset or some sort of known relationship between instances is available. Using this information, models can learn from the full training set even when not all of the instances are labeled. However, all of these methods build on the assumption that labeled and unlabeled sets contain instances of the same classes, which is not the case in NCD.

\textbf{Novelty Detection:}
From a training set with only instances of known classes, the goal of \textit{Novelty Detection} (ND) is to identify if new test observations come from a novel class or not.
This problem is concerned with semantic shift (i.e. the apparition of new classes).
The authors of \cite{oodsurvey} conclude that the goal of ND is only to distinguish novel samples from the training distribution, and not to actually discover the novel classes.
This is the first major difference with Novel Class Discovery, as the ultimate goal of NCD is to explore the novel samples.
The second difference is that in NCD, the known and unknown classes are already split, which is not the case in ND.

\textbf{Novel Class Discovery (NCD):}
In a large part of the literature related to NCD, the goal is to cluster classes in an unlabeled set when a labeled set of known classes is available. Unlike in semi-supervised learning, the classes of these two sets are disjoint. The labeled data is used to reduce the ambiguity that comes with the many potentially valid clustering criteria of the unlabeled data. It is a challenging problem, as the patterns learned from the labeled data may not be relevant or sufficient to cluster the unlabeled data. In \cite{chi2022meta}, the authors explore the assumptions behind NCD and give a formal definition of NCD. They find that this is a theoretically solvable problem under certain assumptions, the most important one being that the labeled and unlabeled data must share good high-level features and that it must be meaningful to separate observations from these two sets.

NCD methods have mainly been developed for the computer vision problem, and the pioneering works involve \textit{Constrained Clustering Network} (CCN) \cite{hsu2018learning}, where the authors approach NCD as a transfer learning problem and rely on the KL-divergence to evaluate the distance between the cluster assignments distributions of two data points. In \textit{Deep Transfer Clustering} (DTC) \cite{han2019learning}, the authors extend the \textit{Deep Embedded Clustering} method \cite{xie2016unsupervised} by taking the available labeled data into account, allowing them to learn more relevant high-level features before clustering the unlabeled data. Another important work is \textit{AutoNovel} \cite{autonovel2021}, where two neural networks update a shared latent space. Finally, \textit{Neighborhood Contrastive Learning} \cite{zhong2021neighborhood}, which is based on \textit{AutoNovel}, adds two contrastive learning terms to the loss to improve the quality of the learned representation. The common denominators of these works are the learning of a latent space and the definition of pairwise pseudo labels for the unlabeled data. Despite showing competitive performance, they are all solving NCD for computer vision problems. Thus, they are able to take advantage of the spatial correlation between the features with specialized architectures and they are not affected by the other challenges of tabular data. Because data augmentation is a crucial component to regularize the network, it is not possible to directly transfer their work on tabular data.



In the next section, the architecture of the proposed method is depicted, along with the two main steps of the training process.


\section{A method for Novel Class Discovery on tabular data}

Given a labeled set $D^l=\{X^l, Y^l\}$ where each instance $x^l \in \mathbb{R}^d$ has a label $y^l \in \{0,1\}^{C^l}$ (representing the one-hot encoding of the $C^l$ classes) and an unlabeled set $D^u=\{X^u\}$, the goal is to identify and predict the classes of $D^u$. In this paper, our setting assumes that the number $C^u$ of classes of $D^u$ is known in advance and that the classes of $D^l$ and $D^u$ are disjoint. Additionally, we follow the formulation of the NCD problem from  \cite{chi2022meta} and suppose that $P_{D^l}(Y|X)$ and $P_{D^u}(Y|X)$ are statistically different and separable, but still share high-level semantic features, so that we can extract general knowledge from $D^l$ of what constitutes a pertinent class. So to \textit{identify} the classes in $D^u$, a transformation $\phi$ must be learned, such that $\phi(X^u)$ is separable.

The proposed method includes two main steps: (i) first, the representation is initialized by pre-training the encoder $\phi$ on $D^l \cup D^u$ without using any label. Then (ii), a supervised classification task and an unsupervised clustering task are jointly solved on the previously learned representation, further updating it. Each of these two steps has its own architecture and training procedure which are described below and in the next sections.

Similarly to \cite{zhong2021neighborhood}, our method is based on AutoNovel \cite{autonovel2021}, and the representation is obtained using a model that includes a feature extractor $\phi(x) = z$, which is a simple combination of multiple dense layers with non-linear activation functions (see Fig.~\ref{fig:tabularncdmodelstep1}).
In the second step, the latent representation of the feature extractor is forwarded to two linear classification and clustering layers followed by Softmax layers (see Fig.~\ref{fig:tabularncdmodelstep3}). 

Despite the fact that for the tasks of classification or regression on tabular data, neither deep nor shallow neural networks have caught up with the performance of classical methods such as XGBoost \cite{xgboost}, LightGBM \cite{LightGBM} or CatBoost \cite{CatBoost}, we believe they are still relevant in our context. Neural networks on tabular data suffer from sparse values, high dimensionality and heterogeneous features. The authors of \cite{deeptabularsurvey} insist that encoders are a good solution to these challenges, as data is projected in a homogeneous latent space of reduced dimensionality. Furthermore, in a well-defined latent space, instances close to each other are likely to belong to the same class. This means that data augmentation and similarity measures relying on this inductive bias can be used with more confidence after projection.

The following subsections describe in more details the two main training steps of the method and the consistency regularization term, which is required in the kind of architecture that is presented here.

\subsection{Initialization of the representation}
\label{sec:ssl}

\begin{figure*}[tb]
	\centerline{\includegraphics[width=0.80\textwidth]{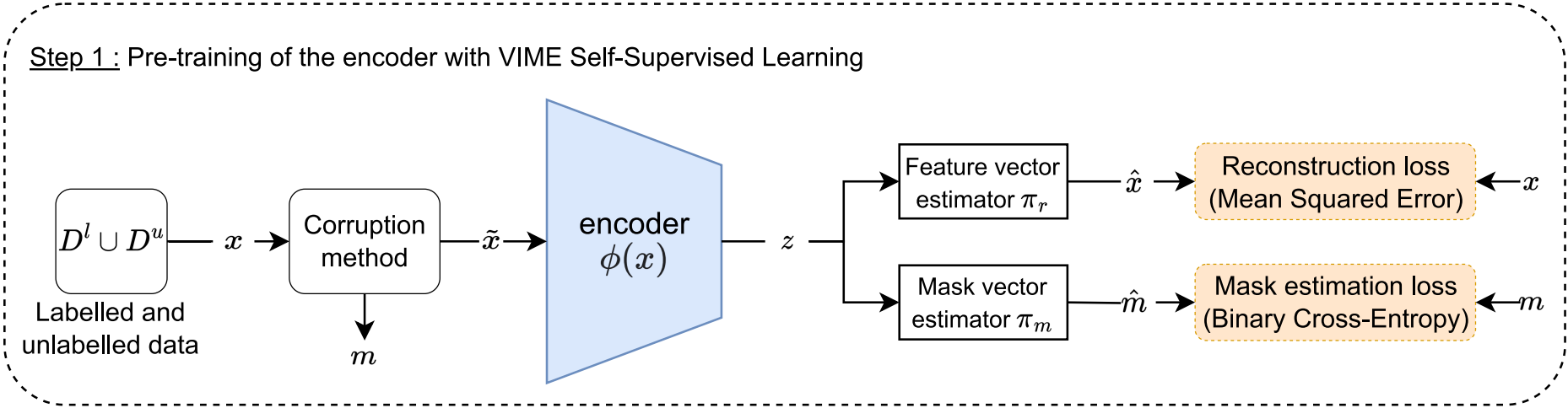}}
    \caption{Architecture of the pre-training step.}
    \label{fig:tabularncdmodelstep1}
\end{figure*}

This first step aims at capturing a common and informative representation of both $D^l$ and $D^u$.
This is important because the representation is used in the next step, among other things, to compute the similarity of pairs of instances and thus determine if examples should belong to the same cluster or not.
Among the possibilities offered in the literature, the way we decided to elaborate this representation is to project examples in a latent space produced by a deep architecture trained on $D^l$ and $D^u$.

To pre-train the latent space using both labeled and unlabeled data, we take advantage of \textit{Self-Supervised Learning} (SSL), and apply the \textit{Value Imputation and Mask Estimation} (VIME) method \cite{NEURIPS2020_7d97667a}.
As the name suggests, VIME defines two pretext tasks to train the encoder $\phi$. From an input vector $x \in \mathbb{R}^d$ that has been corrupted, the objective is to 1) recover the original values and 2) recover the mask used to corrupt the input. The corruption process begins with the generation of a binary mask $m = [m_1, ..., m_d]^\top \in \{0,1\}^d$, with $m_j$ randomly sampled from a Bernoulli distribution with probability $p_m$. The corrupted vector $\Tilde{x}$ is then created by replacing each dimension $j$ of $x$ where $m_j = 1$ with the dimension $j$ of a randomly sampled instance from the training set. So $\Tilde{x} = (1 - m) \odot x + m \odot \bar{x}$, where $\odot$ is the element-wise product and each $\bar{x}_j$ of $\bar{x}$ has been randomly sampled from the empirical marginal distribution of the $j$-th feature of the training set.

The pre-training framework is illustrated in Fig.~\ref{fig:tabularncdmodelstep1}. The encoder and the mask and feature vector estimators are jointly trained with the following optimization problem:
\begin{equation}
    \mathcal{L}_{VIME} = l_{recons.}+ \alpha l_{mask}
    \label{eq:loss_vime}
\end{equation}
where $\alpha$ is a trade-off parameter. Following \cite{NEURIPS2020_7d97667a}, we use $\alpha = 2.0$. The \textit{mean squared error} (MSE) loss is used to optimize the feature vector estimator:
\begin{equation}
    l_{recons.} = \frac{1}{d} \sum_{j=1}^{d}(x_j - \hat{x}_j)^2
    \label{eq:mse_vime}
\end{equation}
where $\hat{x}$ is the input reconstructed from the corrupted vector $\Tilde{x}$. The \textit{binary cross-entropy} (BCE) loss is used to optimize the mask estimator:
\begin{equation}
    l_{mask} = -\frac{1}{d} \sum_{j=1}^{d}[m_j \log(\hat{m}_j) + (1 - m_j)\log(1 - \hat{m}_j)]
    \label{eq:bce_vime}
\end{equation}
where $\hat{m}$ is the estimated mask.

The authors argue that by training an encoder to solve these two tasks, it is possible to capture the correlation between the features and create a latent representation $z$ that contains the necessary information to recover the input $x$. While the general idea of the VIME method is similar to that of a \textit{denoising auto-encoder} (DAE), it reports superior performance in \cite{NEURIPS2020_7d97667a} and there are two major differences. The first is the addition of the mask estimation task, and the second lies in the noise generation process. A DAE will create noisy inputs by adding Gaussian noise or replacing values with zeroes, while VIME randomly selects values from the empirical marginal distribution of each feature. This means that the noisy $\Tilde{x}$ of VIME will be harder to distinguish from the input.

Note that a simple approach to initialize the encoder would be to train a classifier using the known classes. However, the resulting representation would rapidly overfit on these classes and the unique features of the unknown classes would be lost. Instead, by generating pseudo labels from pretext tasks on unlabeled data, a model can be trained to learn informative representations and high level features that are unbiased towards labeled data. This is the general idea behind Self-Supervised Learning and it is a technique commonly used in NCD \cite{cao2021openworld, autonovel2021, zhong2021neighborhood}. An example of it is RotNet \cite{gidaris2018unsupervised}, where the model has to predict the rotation that was applied to an image from the possible 0, 90, 180 and 270 degrees values. Unfortunately, only a few works have attempted to adapt this technique for tabular data \cite{bahri2021scarf, ucar2021subtab, NEURIPS2020_7d97667a} and haven't had the same success.

\subsection{Joint training on the labeled and unlabeled data}
\label{sec:joint}

\begin{figure*}[tb]
	\centerline{\includegraphics[width=0.80\textwidth]{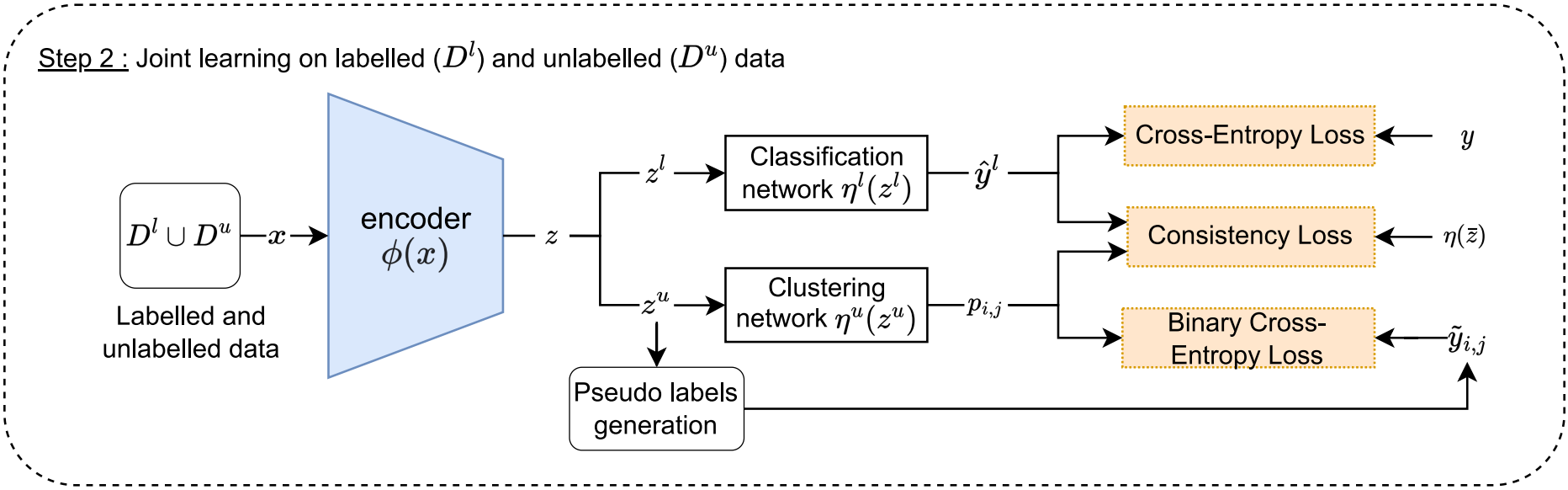}}
    \caption{Architecture of the joint learning step. After the input data $x$ has been projected, labeled data $z^l$ is used to train the classification network $\eta^l$ and unlabeled data is used to train the clustering network $\eta^u$ with the generated pseudo labels as its target.}
    \label{fig:tabularncdmodelstep3}
\end{figure*}

%
%

We formulate the novel class discovery process as multi-task optimization problem. In this step, two new networks are added on top of the previously initialized encoder, each solving different tasks on different data (see Fig.~\ref{fig:tabularncdmodelstep3}). The first is a classification network $\eta^l(z) \in \mathbb{R}^{C^l+1}$ trained to predict 1) the $C^l$ known classes from $D^l$ with the ground-truth labels and 2) a single class formed of the aggregation of the unlabeled data. The second is another classification network trained to predict the $C^u$ novel classes from $D^u$. It will be referred as the \textit{clustering} network $\eta^u(z) \in \mathbb{R}^{C^u}$. These two networks share the same embedding and both update it through back-propagation, sharing knowledge with one another.

The classification network is optimized with the \textit{cross-entropy} loss using the ground-truth one-hot labels $y$:
\begin{equation}
    l_{class.}= - \sum_{c=1}^{C^l+1} y_{c}\log(\eta_c^l(z))
    \label{eq:ce_joint}
\end{equation}
where $z = \phi(x)$. Its role is to guide the representation to include the features of the known classes that are relevant for the supervised classification task.

To train the clustering network $\eta^u$ with unlabeled data in a supervised manner, pseudo labels $\tilde{y}_{i,j} \in \{0, 1\}$ are generated for each pair $(x_i, x_j)$ of unlabeled data in a mini-batch.
This is comparable to a pretext task in Self-Supervised Learning. 
Another possible approach would be to predict the similarity of instances directly, but using pseudo-labels has the advantage of directly associating classes to the instances and does not require an additional clustering of the embedding.
In this case, pseudo labels indicate if a pair of instances should belong to the same class or not, independently of the class number predicted by the network.
They are defined as $\tilde{y}_{i,j} = 1$ if $z_i$ and $z_j$ are similar, and $\tilde{y}_{i,j} = 0$ if they are dissimilar. After the mini-batch $X$ has been projected in the encoder ($Z = \phi(X)$), there are $|Z| - 1$ pairwise pseudo labels defined for each instance $z_i \in Z$.
The clustering network is optimized with the \textit{binary cross-entropy} loss, which is for a single instance $z_i$:
\begin{equation}
    \medmath{
    l_{clust.} = \frac{1}{|Z|-1} \sum\limits_{\substack{j=1 \\ j \neq i}}^{|Z|} \left[ -\tilde{y}_{i,j} \log(p_{i,j}) - (1 - \tilde{y}_{i,j}) \log(1 - p_{i,j}) \right]}
    \label{eq:bce_joint}
\end{equation}
where $p_{i,j} = \eta^u(z_i) \cdot \eta^u(z_j)$ is a score close to 1 if the clustering network predicted the same class for $z_i$ and $z_j$ and close to 0 otherwise. For this reason, $p_{i,j}$ can directly be compared to the pairwise pseudo labels $\tilde{y}_{i,j}$. The intuition is that instances similar to each other in the latent space are likely to belong to the same class. Therefore, $\eta^u$ will create clusters of similar instances, guided by $\eta^l$ with the knowledge of the known classes.

\textbf{Note:} Predicting the unlabeled data as a single new class in the classification network is not done in AutoNovel \cite{autonovel2021} and is one of the main differences with our work. We found that having an overlap of the instances in the classification and clustering networks would improve the performance and result in the definition of cleaner latent space that does not mix labeled and unlabeled data. 


\textbf{Definition of pseudo labels.} Pseudo labels are defined for each unique pair of unlabeled instances in a mini-batch based on similarity. This idea has been employed in many NCD works (e.g. \cite{autonovel2021, hsu2019multiclass, zhong2021neighborhood}), where the most common approach is to define a threshold $\lambda$ for the minimum similarity of pairs of instances to be assigned to the same class.

\begin{figure}[tb]
    \centering
    \subfigure[Representation of the data points of the batch.]{\label{fig:pseudo_labels_1}\includegraphics[width=0.13\textwidth]{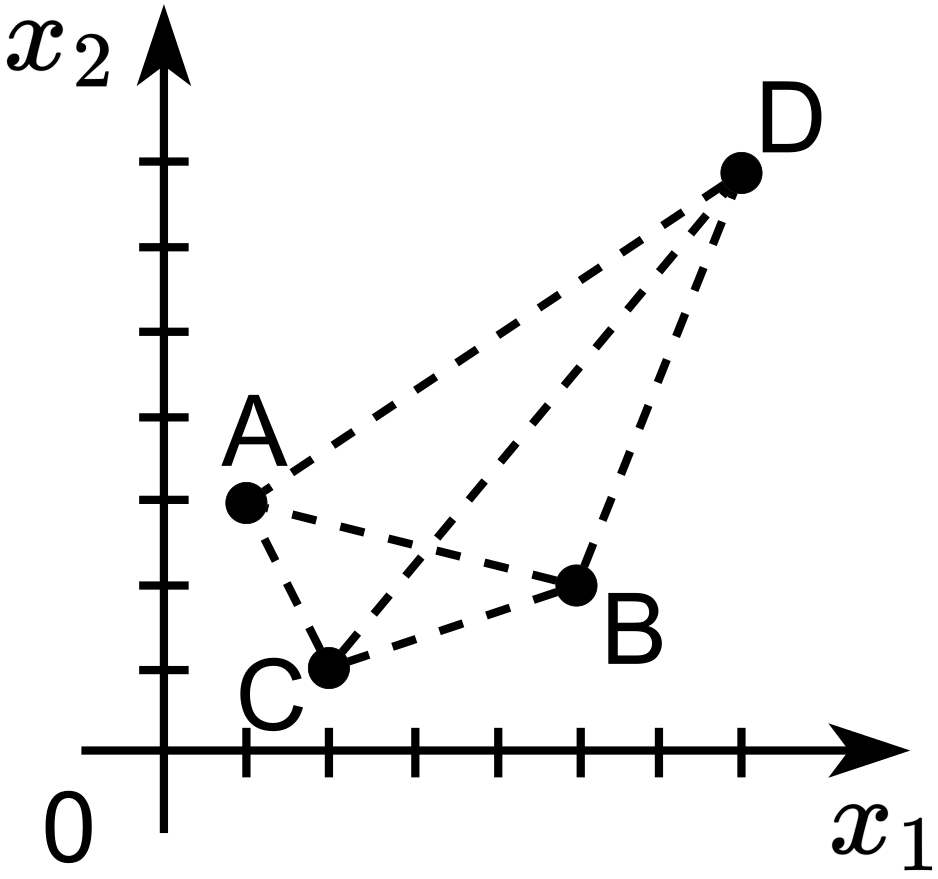}}
    \hfill
    \subfigure[Pairwise similarity matrix.]{\label{fig:pseudo_labels_2}\includegraphics[width=0.13\textwidth]{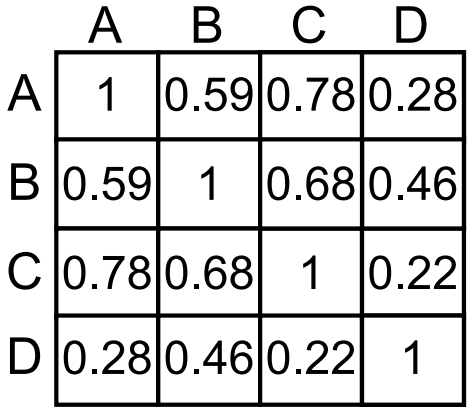}}
    \hfill
    \subfigure[Pairwise pseudo labels matrix for $k = 2$.]{\label{fig:pseudo_labels_3}\includegraphics[width=0.13\textwidth]{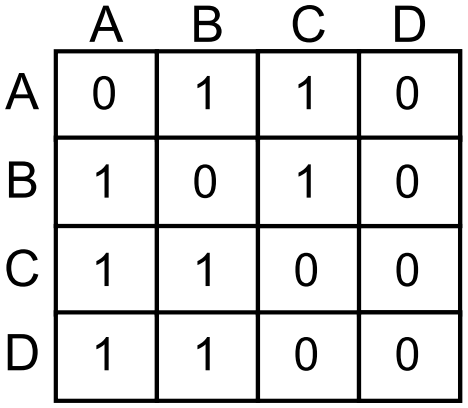}}
    \caption{The pairwise pseudo labels definition process.}
    \label{fig:pseudo_labelling_process}
\end{figure}

However, we found\footnote{Experimentally, see the supplementary material in \url{https://github.com/ColinTr/TabularNCD}} that defining for each point the top $k$ most similar instances as positives was a more reliable method. So, for each pair $(x_i, x_j)$ in the projected batch of unlabeled data $Z$, the pseudo labels are assigned as follows:

\begin{equation}
    \tilde{y}_{i,j} = \mathds{1} [ j \in \underset{\substack{r \in \{ 1, ..., |Z| \} \\ r \neq i}}{\text{argtop}_k } \delta(z_i, z_r) ]
    \label{eq:topk_cos_sim}
\end{equation}

where $\delta(z_i, z_r) = \frac{z_i \cdot z_r}{\lVert z_i \rVert \lVert z_r \rVert}$ is the cosine similarity and $\text{argtop}_k$ is the subset of indices of the $k$ largest elements. This means that each observation has $k$ positive pairs and $|Z|-1-k$ negative pairs. This process is illustrated in Fig.~\ref{fig:pseudo_labelling_process}. From a mini-batch of a few observations (Fig.~\ref{fig:pseudo_labels_1}), the pairwise similarity matrix (Fig.~\ref{fig:pseudo_labels_2}) is derived. Then, the pairwise pseudo labels matrix is defined (Fig.~\ref{fig:pseudo_labels_3}) where in each row, positive relations are set for the $k = 2$ most similar pairs and the rest are negatives. 

\subsection{Consistency regularization}
\label{sec:consistenc_reg}

During the training of the two classification networks, the latent representation of the encoder changes. And since the pseudo labels are defined according to the similarity of instances in the latent space, this causes a \textit{moving target} phenomenon, where the pairwise pseudo labels can differ from one epoch to another. To limit the impact of this problem, a regularization term is needed. Commonly employed in semi-supervised learning \cite{semiSupervEval}, we use ``data augmentation''. The idea is to encourage the model to predict the same class for an instance $x$ and its augmented counterpart $\bar{x}$. But while data augmentation has become a standard in computer vision, the same cannot be said for tabular data. The authors of \cite{deeptabularsurvey} consider the lack of research in tabular data augmentation (along with the difficulty to capture the dependency structure of the data) to be one of the main reasons for the limited success of neural networks in tabular data. Nevertheless, we find that consistency regularization is an essential component for the performance of the method proposed here.

\begin{figure}[tb]
	\centerline{\includegraphics[width=0.12\textwidth]{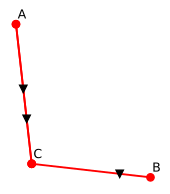}}
    \caption{Illustration of the SMOTE data augmentation technique. The red dots are the batch samples, and the black triangles are the synthetic new samples.}
    \label{fig:smote_like_method_illustration}
\end{figure}

In this paper we use SMOTE-NC (for Synthetic Minority Oversampling Technique \cite{SMOTE}), which is one of the most widely used tabular data augmentation techniques. It synthesizes new samples from the minority classes to reduce imbalance, thus improving the predictive performance of models on these classes. Synthetic samples are generated by taking a point along the line between the considered sample and a random instance among the $k$ closest of the same class (in the original input space). Fig.~\ref{fig:smote_like_method_illustration} illustrates this technique. The larger the $k$ is, the wider and less specific the decision regions of the learned model will be, resulting in better generalization.

This method can be applied directly to instances of the labeled set, however we relax the constraint of selecting same-class instances for the unlabeled set, as no labels are available. It can also be noted that the nearest neighbors for the observations of the unlabeled set are not selected from the labeled set since one of the hypotheses of our problem is that $C^l \cap C^u = \emptyset$ (the same goes for the labeled set).

The \textit{mean squared error} (MSE) is used as the consistency regularization term for both the classification and the clustering network. For the clustering network, it is written as:
\begin{equation}
    l_{reg.} = \frac{1}{C^u} \sum_{c=1}^{C^u} (\eta_c(z) - \eta_c(\bar{z}))^2
    \label{eq:reg}
\end{equation}
where $\bar{z}=\phi(\bar{x})$ is the projection of $x$ augmented with the method inspired from SMOTE-NC. For the classification network, it is averaged over $C^l+1$ instead.


\subsection{Overall loss}
\label{sec:overall_loss}
The method proposed here falls under the domain of Multi-Task Learning (MTL) where instead of focusing on the only task that solves the problem at hand (discovering new classes in $D^u$), additional related tasks are jointly learned (classification of samples in $D^l$). By introducing new inductive biases, the model will prefer some hypotheses over others, guiding the model towards better generalization \cite{multitask}. Our architecture falls under the \textit{hard parameter sharing} domain of MTL, where the first layers are shared between the different tasks (i.e. the encoder), and the last few layers are task-specific.


We chose to adopt an alternating optimization strategy, which was introduced in \cite{2109.11678}. In this case, we define one objective function and one individual optimizer per task. The loss of the classification network is:
\begin{equation}
    \mathcal{L}_{classification} = w_1 l_{classif.}+ (1- w_1) l_{reg.}
    \label{eq:l_classification}
\end{equation}
And the loss of the clustering network can be written as:
\begin{equation}
    \mathcal{L}_{clustering} = w_2 l_{clust.} + (1 - w_2) l_{reg.}
    \label{eq:l_clustering}
\end{equation}
where $w_1$ and $w_2$ are trade-off hyper-parameters for balancing the weight of the consistency regularization terms.

\textbf{Description of the training process: }
The classification network and the clustering network are trained alternately. For each mini-batch during training, the classification network and the encoder are first updated through back-propagation with the loss from \eqref{eq:l_classification}. Then, the unlabeled data of the same mini-batch is forwarded once again through the encoder to compute the loss from \eqref{eq:l_clustering}, which is back-propagated to update the clustering network and the encoder once more.

\section{Experiments}

\subsection{Datasets and experimental details}
\label{sec:experimental_details}

\textbf{Datasets.} To evaluate the performances of the method proposed here, a variety of datasets in terms of application domains and characteristics have been chosen (see Table~\ref{table:datasets_description}). Six public tabular classification datasets were selected: Forest Cover Type \cite{forestcover}, Letter Recognition \cite{letterrecog}, Human Activity Recognition \cite{humanactivity}, Satimage \cite{uci}, Pen-Based Handwritten Digits Recognition\cite{uci} and 1990 US Census Data \cite{uci}, as well as MNIST \cite{deng2012mnist}, which was flattened to transform the $28 \times 28$ grayscale images into vectors of $784$ attributes.
We pre-process the numerical features of all the datasets to have zero-mean and unit-variance, while the categorical features are one-hot encoded.
If the training and testing data were not already split, $70\%$ were kept for training while the remaining $30\%$ were used for testing. 

Following the same procedure found in Novel Class Discovery articles \cite{cao2021openworld, autonovel2021, zhong2021neighborhood}, we hide the labels of some classes to create the \textit{unknown} classes. And the capacity of the compared methods to recover these classes is then evaluated. Around 50\% of the classes are hidden, resulting in further splitting of the training and testing sets into labeled and unlabeled subsets. The partitions of the 7 datasets can be found in Table \ref{table:datasets_description}.

\begin{table}[tb]
    \fontsize{7}{8}\selectfont
    \caption{Statistical information of the selected datasets.}
    \centering
    \setlength{\tabcolsep}{2pt}
        \begin{tabular}{l|c|c|c|c|c|c}
            \hline
            \multirow{2}{*}{Dataset name} & Attri- & \# classes & \# train  & \# train  & \# test   & \# test  \\
              & butes & $C^l$ / $C^u$ & labeled & unlabeled & labeled  &   unlabeled \\
            \hline
            MNIST \cite{deng2012mnist}            & 784 & 5 / 5  & 30,596 & 29,404 & 5,139  & 4,861  \\
            Forest Cover type \cite{forestcover}  & 54  & 4 / 3  & 6,480  & 4,860  & 36,568 & 13,432 \\
            Letter recognition \cite{letterrecog} & 16  & 19 / 7 & 10,229 & 3,770  & 4,296  & 1,704  \\
            Human activity \cite{humanactivity}   & 562 & 3 / 3  & 3,733  & 3,619  & 1,494  & 1,453  \\
            Satimage \cite{uci}                   & 36  & 3 / 3  & 2,525  & 1,976  & 1,042  & 887    \\
            Pendigits \cite{uci}                  & 16  & 5 / 5  & 3,777  & 3,717  & 1,764  & 1,734  \\
            1990 US Census \cite{uci}             & 67  & 12 / 6 & 50,000 & 50,000 & 31,343 & 18,657 \\
            \hline
        \end{tabular}
        \label{table:datasets_description}
\end{table}

\textbf{Evaluation metrics.} To thoroughly assess the performance of the methods compared in the experiments described below, we report four different metrics. The first two are the clustering accuracy (ACC) and balanced accuracy (BACC), which can be computed after the optimal linear assignment of the class labels is solved with the Hungarian algorithm \cite{Kuhn55hungarian}. The use of the balanced accuracy is justified by the unbalanced class distribution of some datasets, which resulted in inflated accuracy scores that were not truly representative of the performance of the method.
Another reported metric is the Normalized Mutual Information (NMI). It measures the correspondence between two sets of label assignments and is invariant to permutations. 
%
%
Finally, we compute the Adjusted Rand Index (ARI). It measures the similarity between two clustering assignments.%

These four metrics range between 0 and 1, with values closer to 1 indicating a better agreement to the ground truth labels. The ARI is an exception and can yield values in $[-1,+1]$, with negative values when the index is less than the $Expected\_RI$. The metrics reported in the next section are all computed on the unlabeled instances from the test sets.

\textbf{Competing methods.} As expressed in Section \ref{sec:intro}, to the best of our knowledge, there is no other method that solves the particular Novel Class Discovery problem for tabular datasets. Nonetheless, we can compare ourselves to unsupervised clustering methods. This will also allow us to show that our method provides an effective way of incorporating knowledge from known classes. We choose the $k$-means algorithm for its simplicity and wide adoption, as well as the Spectral Clustering \cite{Luxburg07} method for its known good results to discover new patterns as in Active Learning \cite{wang2010active}. We set $k$ to the ground truth (i.e. $k = C^u$) for both clustering methods, as it was already assumed that it is known in the proposed method.

We also define a simple baseline that clusters unlabeled data while still capitalizing on known classes: (i) first, a usual classification neural network is trained on the known classes from $D^l$; (ii) then the classifier's penultimate layer is used as a feature embedding for $D^u$. In this projection, a $k$-means is applied to assign labels to the unlabeled test instances. Compared to the unsupervised approaches, this technique has the advantage of learning the patterns of the known classes in a homogeneous latent space. 
However, we still expect it to perform sub-optimally as the unique features and patterns of the unlabeled data might be lost in the last hidden layer after training.



\textbf{Implementation details.} For all datasets, we employ an encoder composed of 2 dense hidden layers that keeps the dimension of the input with activation functions and dropout values that are optimized hyper-parameters. Following VIME \cite{NEURIPS2020_7d97667a} and AutoNovel \cite{autonovel2021}, we use a single linear layer for the mask estimator, vector estimator, classification network and clustering network.

The hyper-parameters (see Table \ref{table:hyperparameters3} in appendix) of the proposed method are optimized with a Bayesian search on a validation set which represents $20\%$ of the training set. During the Self-Supervised Learning step (see Section \ref{sec:ssl}), we observe little impact on the final performance of the model when optimizing the hyper-parameters specific to the VIME method, and therefore use the values suggested in the original paper: a learning rate of 0.001, a corruption rate of 30\% and a batch size of 128. In the joint training step (see Section \ref{sec:joint}), we use a larger batch size of 512 as the observations are randomly sampled from the merged labeled and unlabeled data. The hyper-parameters of this step are the learning rates of both classification and clustering network optimizers, along with the trade-off parameters of their respective losses. The $top$ $k$ number of positive pairs defined by \eqref{eq:topk_cos_sim} and the $k$ $neighbors$ considered in the data augmentation method of Sec. \ref{sec:consistenc_reg} are also optimized and can all be found in Table \ref{table:hyperparameters3} in appendix. We use AdamW \cite{adamw} as the optimizer for the pre-training step and for the two optimizers of the joint training step, and find that 30 epochs are enough to converge for both steps on any of the tabular datasets used.

We implemented our method under Python 3.7.9 and with the PyTorch 1.10.0 \cite{NEURIPS2019_9015} library. The experiments were conducted on Nvidia 2080 Ti and Nvidia Quadro T1000 GPUs. The networks are initialized with random weights, and following \cite{autonovel2021}, the results are averaged over 10 runs. The code is publicly available at \url{https://github.com/ColinTr/TabularNCD}.

\subsection{Results}

\begin{table}[tb]
    \caption{Performance of TabularNCD on the unknown classes.}
    \begin{center}
        \setlength{\tabcolsep}{3pt}
        \begin{tabular}{l | l c c c c}
            \hline
            Dataset & Method & BACC (\%) & ACC (\%) & NMI & ARI \\
            \hline
            \multirow{4}{*}{MNIST}      & Baseline         & 57.7$\pm$4.7          & 57.6$\pm$4.5          & 0.37$\pm$0.2           & 0.31$\pm$0.3           \\
                                        & Spect. clust     & -                     & -                     & -                      & -                      \\
                                        & \textit{k}-means & 60.1$\pm$0.0          & 61.1$\pm$0.0          & 0.48$\pm$0.0           & 0.38$\pm$0.0           \\
                                        & TabularNCD       & \textbf{91.5$\pm$4.1} & \textbf{91.4$\pm$4.2} & \textbf{0.82$\pm$0.06} & \textbf{0.81$\pm$0.04} \\
            \hline
            \multirow{4}{*}{Forest}     & Baseline         & 55.6$\pm$2.0          & 68.5$\pm$1.4          & 0.27$\pm$0.02          & 0.15$\pm$0.01          \\
                                        & Spect. clust     & 32.1$\pm$1.4          & 85.8$\pm$4.0          & 0.01$\pm$0.01          & 0.09$\pm$0.01          \\
                                        & \textit{k}-means & 32.9$\pm$0.0          & 62.0$\pm$0.0          & 0.04$\pm$0.00          & 0.05$\pm$0.00          \\
                                        & TabularNCD       & \textbf{66.8$\pm$0.6} & \textbf{92.2$\pm$0.2} & \textbf{0.37$\pm$0.09} & \textbf{0.56$\pm$0.09} \\
            \hline
            \multirow{4}{*}{Letter}     & Baseline         & 55.7$\pm$3.6          & 55.9$\pm$3.6          & 0.49$\pm$0.04          & 0.33$\pm$0.04          \\
                                        & Spect. clust     & 45.3$\pm$4.0          & 45.3$\pm$4.0          & 0.48$\pm$0.03          & 0.18$\pm$0.03          \\
                                        & \textit{k}-means & 50.2$\pm$0.6          & 49.9$\pm$0.6          & 0.40$\pm$0.01          & 0.28$\pm$0.01          \\
                                        & TabularNCD       & \textbf{71.8$\pm$4.5} & \textbf{71.8$\pm$4.5} & \textbf{0.60$\pm$0.04} & \textbf{0.54$\pm$0.04} \\
            \hline
            \multirow{4}{*}{Human}      & Baseline         & 80.0$\pm$0.5          & 78.0$\pm$0.6          & 0.64$\pm$0.01          & 0.62$\pm$0.01          \\
                                        & Spect. clust     & 70.2$\pm$0.0          & 69.4$\pm$0.0          & 0.72$\pm$0.00          & 0.60$\pm$0.00          \\
                                        & \textit{k}-means & 75.3$\pm$0.0          & 77.0$\pm$0.0          & 0.62$\pm$0.00          & 0.59$\pm$0.00          \\
                                        & TabularNCD       & \textbf{98.9$\pm$0.2} & \textbf{99.0$\pm$0.2} & \textbf{0.95$\pm$0.01} & \textbf{0.97$\pm$0.01} \\
            \hline
            \multirow{4}{*}{Satimage}   & Baseline         & 53.8$\pm$3.4          & 53.9$\pm$4.2          & 0.25$\pm$0.03          & 0.22$\pm$0.03          \\
                                        & Spect. clust     & 82.2$\pm$0.1          & 77.8$\pm$0.1          & 0.51$\pm$0.00          & 0.46$\pm$0.00          \\
                                        & \textit{k}-means & 73.7$\pm$0.3          & 69.2$\pm$0.2          & 0.30$\pm$0.00          & 0.28$\pm$0.00          \\
                                        & TabularNCD       & \textbf{90.8$\pm$4.0} & \textbf{91.4$\pm$5.0} & \textbf{0.71$\pm$0.11} & \textbf{0.79$\pm$0.07} \\
            \hline
            \multirow{4}{*}{Pendigits}  & Baseline         & 72.8$\pm$5.5          & 72.8$\pm$5.4          & 0.62$\pm$0.06          & 0.54$\pm$0.07          \\
                                        & Spect. clust     & {84.0$\pm$0.0}        & {84.0$\pm$0.0}        & \textbf{0.78$\pm$0.00} & 0.67$\pm$0.00          \\
                                        & \textit{k}-means & 82.5$\pm$0.0          & 82.5$\pm$0.0          & 0.72$\pm$0.00          & 0.63$\pm$0.00          \\
                                        & TabularNCD       & \textbf{85.5$\pm$0.7} & \textbf{85.6$\pm$0.8} & 0.76$\pm$0.02          & \textbf{0.71$\pm$0.02} \\
            \hline
            \multirow{4}{*}{Census}     & Baseline         & 53.0$\pm$3.5          & \textbf{55.0$\pm$6.5} & 0.49$\pm$0.02          & 0.30$\pm$0.03          \\
                                        & Spect. clust     & 23.6$\pm$3.3          & 51.3$\pm$5.5          & 0.24$\pm$0.11          & 0.18$\pm$0.09          \\
                                        & \textit{k}-means & 38.5$\pm$2.6          & 49.8$\pm$3.6          & 0.41$\pm$0.05          & 0.28$\pm$0.03          \\
                                        & TabularNCD       & \textbf{61.9$\pm$0.6} & 50.1$\pm$0.9          & 0.48$\pm$0.01          & 0.30$\pm$0.00          \\
            \hline
        \end{tabular}
        \label{table:tabularncd_results}
    \end{center}
    \footnotesize{The standard deviation is computed over 10 executions. The 2 unsupervised clustering methods (Spect. clust and $k$-means) are only fitted to the test instances belonging to the unknown classes. Values for the spectral clustering of MNIST are missing as the execution did not complete under 1 hour.}
\end{table}

\textbf{Comparison to the competing methods.} In Table \ref{table:tabularncd_results}, we report the performance of the 4 competing methods and on the 7 datasets for the clustering task. The results show that TabularNCD achieves higher performance than the baseline and the two unsupervised clustering methods, on all considered datasets and on all metrics. The improvements in accuracy over competing methods ranges between 1.6\% and 21.0\%. This proves that even for tabular data, useful knowledge can be extracted from already discovered classes to guide the novel class discovery process. 

\begin{figure}[tb]
    \centerline{\includegraphics[width=0.35\textwidth]{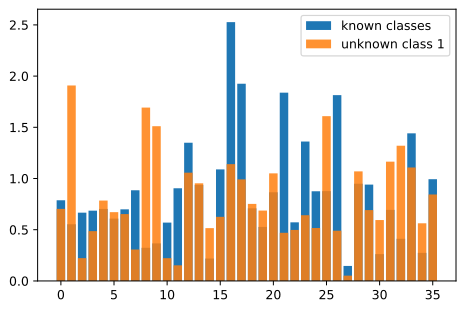}}
    \caption{Per-attribute coefficients of a logistic regression trained to predict the known and unknown classes. The dataset is \textit{Satimage}.} 
    \label{fig:feature_importance}
\end{figure}

While the baseline competitor ($k$-means on a projection learned on the labeled data, see Sec.~\ref{sec:experimental_details}) improves the performance of $k$-means for some datasets (notably the Census and Forest datasets), it still lags behind our method in general. The baseline obtains a lower score than the simple $k$-means on only 3 datasets, which means that in some cases, the features extracted from the known classes are insufficient to discriminate novel unseen data. This is the case for the \textit{Satimage} dataset, where the performance of the baseline is much lower compared to the simple $k$-means on the original data. To understand this phenomenon, we compare the importance of the features used to predict the known classes to the importance of the features of the unknown classes. Because the features have been standardized, we can simply look at the coefficients of a logistic regression. Fig.~\ref{fig:feature_importance} illustrates this experiment: the 5 most important coefficients in the prediction of one of the unknown classes are more than twice as large as the addition of the coefficients of the attributes used to predict the known classes. Because these attributes that are critical in the discrimination of this unknown class are relatively unimportant in the discrimination of the known classes, they are lost during the training of the classifier that serves as a feature extractor in the baseline, which results in a projection of the unlabeled instances of poor quality. This issue is more pronounced as the known and unknown classes are more dissimilar. The latent representation of TabularNCD is pre-trained using SSL on all the available data, which explains why it is unaffected by the phenomenon described above.

\textbf{Comparison to a NCD method on MNIST.} CD-KNet-Exp \cite{wang2020openworld} (for \textit{Class Discovery Kernel Networks with Expansion}) is one of the first methods that attempts to solve the NCD problem. They define their problem as an Open World Class Discovery problem, which is actually another name for Novel Class Discovery. In their proposed method, they start by training a classifier on the known classes of $D^l$, and then fine-tune it on both $D^l$ and $D^u$ before applying a $k$-means on the learned latent representation to get the cluster assignments. Similarly to our experimental protocol, they select the first 5 digits of MNIST as known classes and use the last 5 as new unlabeled classes to discover. For this reason, we can directly compare their reported performance to ours in Table \ref{table:cdknet}. From this table, we see that in terms of clustering accuracy, our method performs 4.5\% better than their simpler implementation CD-KNet, and 3.1\% worse than their complete approach CD-KNet-Exp, where the previously learned classifier is re-trained with the pseudo labels assigned using $k$-means. Obtaining comparable results on MNIST is very encouraging because unlike our approach, CD-KNet-Exp takes advantage of convolutional neural networks as they only use image data. Their re-training technique could also be explored in future work to improve performance.

\begin{table}[tb]
    \caption{Performance of the proposed method against CD-KNet \cite{wang2020openworld}.}
    \centering
        \begin{tabular}{l | c c c}
            \hline
            Dataset & \multicolumn{3}{c}{MNIST} \\
            \hline
            Method           & ACC (\%)           & NMI            & ARI            \\
            \hline
            TabularNCD       & 91.4          & 0.823          & 0.812          \\
            CD-KNet          & 86.9          & 0.683          & 0.707          \\
            CD-KNet-Exp      & \textbf{94.5} & \textbf{0.856} & \textbf{0.869} \\
            \hline
        \end{tabular}
        \label{table:cdknet}
\end{table}

\textbf{Ablation study.} The usefulness of each of the components of the proposed method is estimated in Table~\ref{table:ablation_study} by ablating them and comparing the metrics after training to the metrics of full method. The first observation that can be made is that each component has a positive influence, and removing any of them results in a drop in performance. The most important one is the BCE \eqref{eq:bce_joint}, since without it the clustering network degenerates to a trivial solution where it predicts the same class for all observations. Next is the MSE from the consistency loss \eqref{eq:reg}. The substantial drop in performance was expected as its role is crucial to (1) reduce the moving target phenomenon introduced in Sec.~\ref{sec:consistenc_reg} and (2) regularize the network to improve its generalization. We also observe an important decrease in performance when removing the CE \eqref{eq:bce_joint}, meaning that in the case of the Satimage dataset, jointly learning a classification network with the clustering network helps guide the clustering process as intended. Finally, while it is beneficial to pre-train the encoder with SSL, the impact is small. This can be explained in part by the limited success that SSL works have had in the challenging area of tabular data.
\begin{table}[!hbt]
    \caption{Ablation study of the proposed method.}
        \centering
        \begin{tabular}{ l | c  c  c  c }
            \hline
            Method               & BACC (\%) & ACC (\%) & NMI & ARI \\
            \hline
            TabularNCD           & 90.8$\pm$4.0 & 91.4$\pm$5.0 & 0.71$\pm$0.11 & 0.79$\pm$0.07 \\
            \hline
            - w/o SSL & 88.4$\pm$5.3 & 88.6$\pm$7.0 & 0.67$\pm$0.15 & 0.67$\pm$0.10 \\
            - w/o CE  & 72.0$\pm$6.1 & 69.5$\pm$6.0 & 0.44$\pm$0.12 & 0.49$\pm$0.08 \\
            - w/o BCE & 33.3$\pm$0.0 & 51.7$\pm$0.0 & 0.00$\pm$0.00 & 0.00$\pm$0.00 \\
            - w/o MSE & 66.7$\pm$5.7 & 63.9$\pm$4.4 & 0.44$\pm$0.02 & 0.37$\pm$0.02 \\  \hline 
        \end{tabular}
        \label{table:ablation_study}
    \footnotesize{\textbf{SSL:} Self-Supervised Learning, \textbf{CE:} Cross Entropy loss of the classification network, \textbf{BCE:} Binary Cross Entropy loss of the clustering network, \textbf{MSE:} Mean Squared Error consistency loss. The dataset is Satimage \cite{uci}.}
\end{table}

\begin{figure*}[tb]
	\begin{center}
		\subfigure[Original data]{\label{fig:latent_init}\includegraphics[width=0.24\textwidth]{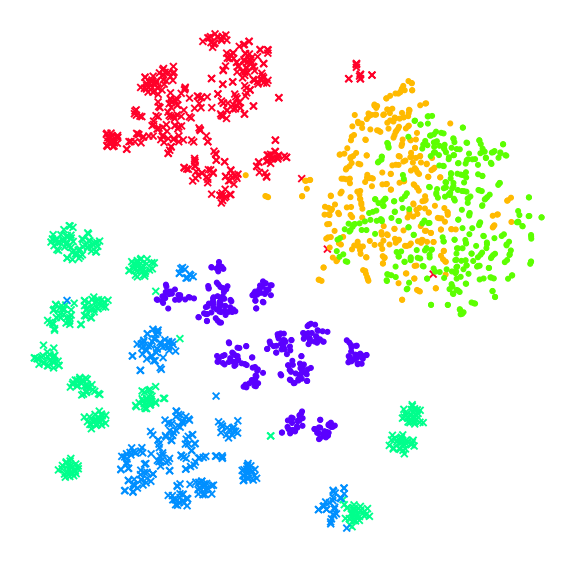}}
		\subfigure[After SSL]{\label{fig:latent_ssl}\includegraphics[width=0.24\textwidth]{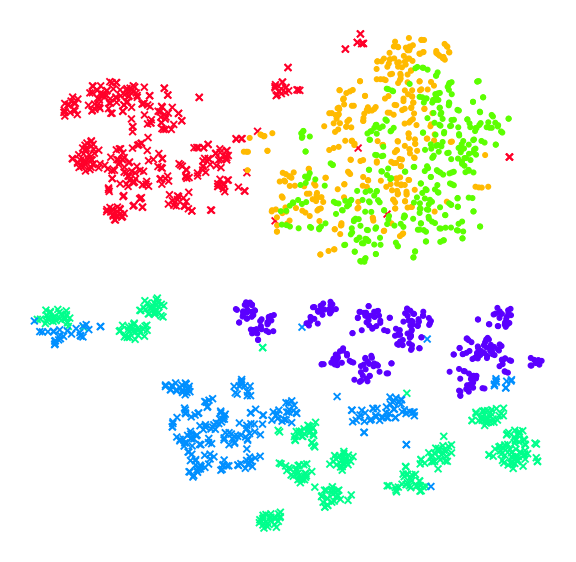}}
		\subfigure[Joint training epoch 15]{\label{fig:latent_15}\includegraphics[width=0.24\textwidth]{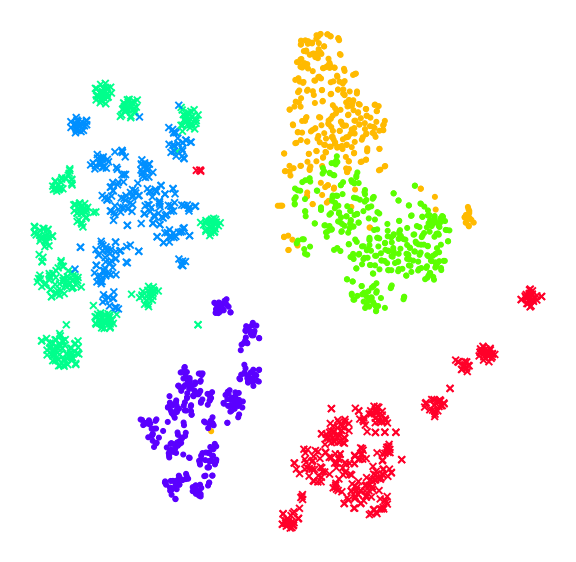}}
		\subfigure[Joint training epoch 30]{\label{fig:latent_30}\includegraphics[width=0.24\textwidth]{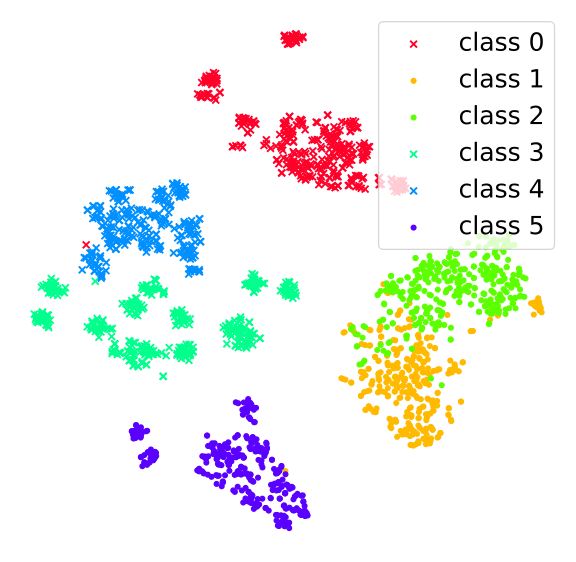}}
        \caption{Evolution of the t-SNE during the joint training of the model on the \textit{Human Activity Recognition} dataset.}
        \label{fig:latent_space}
	\end{center}
\end{figure*}

\textbf{Visualization.} In addition to quantitative results, we also report a qualitative analysis showing the feature space that is learned. In Fig.~\ref{fig:latent_space}, we visualize the evolution of the representation of all the classes during the training process of the proposed method. Fig~\ref{fig:latent_init} corresponds to the original data, while the next figures are the data projected in the latent space. Classes overlap in the original representation, as well as in the latent space after Self-Supervised Learning (Fig~\ref{fig:latent_ssl}). However, the SSL step initialized the encoder to a reasonable representation. This means that at the beginning of the joint training, the pseudo labels defined on the unlabeled data using \eqref{eq:topk_cos_sim} will be accurate. During the joint training in the Figures \ref{fig:latent_15} and \ref{fig:latent_30}, the classes are more and more separated, showing that the proposed method is able to successfully discover novel classes using knowledge from known classes. The plot clearly shows that our model produces feature representations where samples of the same class are tightly grouped.



\section{Conclusions and future work}
In this paper, we have proposed a first solution to the Novel Class Discovery problem in the challenging environment of tabular data. We demonstrated the effectiveness of our proposed approach, TabularNCD, through extensive experiments and careful analysis on 7 public datasets against unsupervised clustering methods. The greater performance of our method has shown that it is possible to extract knowledge from already discovered classes to guide the discovery process of novel classes, which demonstrates that NCD is not only applicable to images but also on tabular data. Lastly, the original method of defining pseudo labels proposed here has proven to be reliable even in the presence of unbalanced classes.

The recent advances of deep learning in tabular data are worth investigating in future work. Specifically, we will explore Generative Adversarial Networks and Variational Auto-Encoders as substitutes for the model's current simple encoder, which is a core component of our method. Furthermore, the assumption that the number of novel classes is known is a limitation of our method, and will certainly be a future direction of our work.

\bibliographystyle{IEEEtran}
\bibliography{main}


\begin{appendices}



\section*{Hyper-parameters influence.}
\begin{figure}[htb]
	\begin{center}
		\subfigure[Pseudo label $top\_k$ (in \%)]{\label{fig:perf_evol_top_k}\includegraphics[height=25mm]{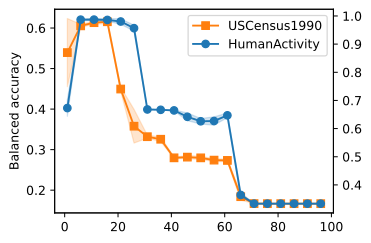}}
		\hfill
		\subfigure[Data aug. $k\_neighbors$]{\label{fig:perf_evol_k_neighbors}\includegraphics[height=25mm]{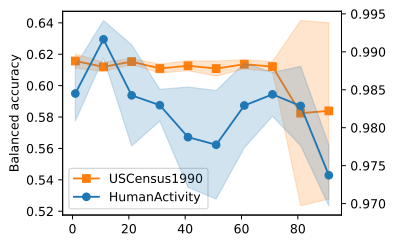}}
		\hfill
		\subfigure[Trade-off $w1$]{\label{fig:perf_evol_w1}\includegraphics[height=25mm]{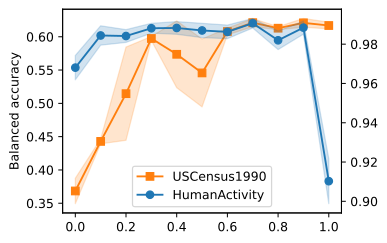}}
		\hfill
		\subfigure[Trade-off $w2$]{\label{fig:perf_evol_w2}\includegraphics[height=25mm]{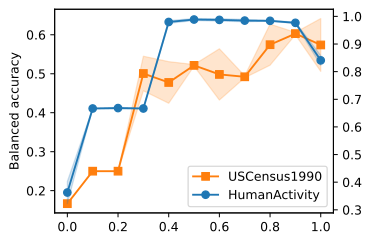}}
        \caption{Visualization of the influence of each parameter on the performance on the Human Activity Recognition and  1990 US Census datasets.}
        \label{fig:parameter_influence}
	\end{center}
\end{figure}

The method proposed here has a few hyper-parameters that must be tuned to get optimal performance. In Fig~\ref{fig:parameter_influence}, we study some of the most important hyper-parameters. The first is the $top\_k$ from \eqref{eq:topk_cos_sim}, it represents the percentage of pairs that will be regarded as positives for each instance. It varies in the range [1, 100] in Fig.~\ref{fig:perf_evol_top_k}, with ideal values between 6 and 16\% for the two datasets considered. The performance is shaped in ``steps'', decreasing as $top\_k$ increases. This shows that after certain thresholds the model merges some of the novel classes together, with values superior to 70\% causing the model to predict a single class for all observations.

Then, we vary the $k\_neighbors$ in [1, 90]. This parameter controls the number of closest neighbors considered in the data augmentation (see Sec.~\ref{sec:consistenc_reg}). Fig.~\ref{fig:perf_evol_k_neighbors} shows under 70, all values produce good results. The \textit{Human} datasets appears to have high variance, but this is simply a result of the very small range of the performance (between 0.970 and 0.995).

Finally, we study $w1$ and $w2$ in Figs.~\ref{fig:perf_evol_w1} and \ref{fig:perf_evol_w2} by varying them in [0, 1]. They control the trade-off between the consistency loss and loss of the classification or clustering networks (see Sec.\ref{sec:overall_loss}). Because the classification task is secondary to the clustering problem, $w1$ has less influence on performance, and good values range between 0.3 and 0.9 for both datasets. On the other hand, $w2$ is very important for the accuracy of the method, and good values range between 0.8 and 0.9 for the \textit{Census} dataset and between 0.4 and 0.9 for the \textit{Human} dataset.

\section*{Influence of data representation for the k-means}
In table \ref{table:k_means_different_latent}, the quality of different data representations of MNIST are compared using $k$-means. We observe that after Self-Supervised Learning, the encoder learned a representation of the original data where the $k$-means algorithm performs slightly better than on the original data ($+2.8\%$ in accuracy). And after joint training, the unknown classes are even more cleanly separated, as $k$-means gains $+24.3\%$ in accuracy over the original data representation.
\begin{table}[!h]
    \caption{Clustering performance on new classes of $k$-means on different data representations.}
    \begin{center}
        \begin{tabular}{l | c c c c}
            \hline
            Method & \multicolumn{4}{c}{$k$-means on MNIST} \\
            \hline
            Data representation & BACC (\%) & ACC (\%) & NMI & ARI \\
            \hline
            (1) Original data   & 60.1 & 61.1 & 0.48 & 0.38 \\
            (2) Latent w/ SSL   & 63.8 & 63.9 & 0.501 & 0.414 \\
            (3) Latent w/ joint & 84.9 & 85.4 & 0.748 & 0.728 \\
            \hline
            \end{tabular}
        \label{table:k_means_different_latent}
    \end{center}
    \footnotesize{Performance of the $k$-means algorithm on different data representation of the MNIST dataset. (1) is the original data, (2) is the data projected by an encoder trained with Self-Supervised Learning (Sec.~\ref{sec:ssl}), and (3) is the data projected by an encoder jointly trained by the proposed method (Sec.~\ref{sec:joint})}
\end{table}

\section*{Hyper-parameters values}
\label{sec:hyperparams}
We optimize the most important hyper-parameters for each dataset and report their values in table \ref{table:hyperparameters3}. The constants are:
\textit{batch size = 512},
\textit{encoder layers sizes = [d, d, d]},
\textit{$\alpha$ = 2.0},
\textit{ssl lr = 0.001},
\textit{epochs = 30} and
\textit{$p\_m$ = 0.30}.

\begin{table}[!htb]
    \fontsize{7}{8}\selectfont
    \caption{Best hyper-parameters found.}
    \begin{center}
    \setlength{\tabcolsep}{1pt}
        \begin{tabular}{ l | c | c | c | c | c | c | c }
            \hline
            Parameter & MNIST & Forest & Letter & Human & Satimage & Pendigits & Census \\
            \hline
            cosine topk     & 15.015    & 19.300   & 2.019    & 15.277    & 6.214    & 5.609    & 13.96     \\
            w1              & 0.8709    & 0.1507   & 0.4887   & 0.4560    & 0.80     & 0.7970   & 0.8104    \\
            w2              & 0.6980    & 0.8303   & 0.9350   & 0.6335    & 0.8142   & 0.8000   & 0.9628    \\
            $lr_{classif.}$ & 0.009484  & 0.001876 & 0.009906 & 0.001761  & 0.007389 & 0.006359 & 0.005484  \\
            $lr_{cluster.}$ & 0.0009516 & 0.007191 & 0.007467 & 0.001017  & 0.008819 & 0.009585 & 0.0008563 \\
            k neighbors     & 4         & 9        & 6        & 15        & 11       & 10       & 7         \\
            dropout         & 0.01107   & 0.09115  & 0.07537  & 0.2959    & 0.4210   & 0.01652  & 0.06973   \\
            activation fct. & Sigmoid   & Sigmoid  & ReLU     & ReLU      & ReLU     & ReLU     & ReLU      \\
            \hline
        \end{tabular}
    \label{table:hyperparameters3}
    \end{center}
    \footnotesize{``cosine topk'' is here a percentage of the max number of pairs of instances in a batch.}
\end{table}


\end{appendices}

\end{document}